\documentclass{article}
\usepackage[numbers]{natbib}

\usepackage{arxiv}

\usepackage[utf8]{inputenc} 
\usepackage[T1]{fontenc}    
\usepackage{hyperref}       
\usepackage{url}            
\usepackage{booktabs}       
\usepackage{amsfonts}       
\usepackage{amsmath}        
\usepackage{nicefrac}       
\usepackage{microtype}      
\usepackage{lipsum}         
\usepackage{graphicx}
\usepackage{natbib}
\usepackage{doi}

\title{ELSA: A Style-Aligned Dataset for Emotionally Intelligent Language Generation}

\author{ {\hspace{1mm}Vishal Gandhi} \\
	Joyspace AI\\
	\texttt{vishal@joyspace.ai} \\
	\And
	{\hspace{1mm}Sagar Gandhi} \\
	Joyspace AI\\
	\texttt{sagar@joyspace.ai} \\
}

\renewcommand{\headeright}{} 
\renewcommand{\undertitle}{}
\renewcommand{\shorttitle}{}

\begin{document}
\maketitle
\pagestyle{plain} 

\begin{abstract}
Advancements in emotion-aware language processing increasingly shape vital NLP applications ranging from conversational AI and affective computing to computational psychology and creative content generation. Existing emotion datasets either lack emotional granularity or fail to capture necessary stylistic diversity, limiting the advancement of effective emotion-conditioned text generation systems. Seeking to bridge this crucial gap between granularity and style diversity, this paper introduces a novel systematically constructed dataset named ELSA (Emotion and Language Style Alignment Dataset)\footnote{\url{https://huggingface.co/datasets/joyspace-ai/ELSA-Emotion-and-Language-Style-Alignment-Dataset}} leveraging fine-grained emotion taxonomies adapted from existing sources (dair-ai/emotion dataset and GoEmotions taxonomy). This dataset comprises multiple emotionally nuanced variations of original sentences regenerated across distinct contextual styles (conversational, formal, poetic, and narrative) using advanced Large Language Models (LLMs). Rigorous computational evaluation using metrics such as perplexity, embedding variance, readability, lexical diversity, and semantic coherence measures validates the dataset’s emotional authenticity, linguistic fluency, and textual diversity. Comprehensive metric analyses affirm its potential to support deeper explorations into emotion-conditioned style-adaptive text generation. By enabling precision-tuned emotionally nuanced language modeling, our dataset creates fertile ground for research on fine-grained emotional control, prompt-driven explanation, interpretability, and style-adaptive expressive language generation with LLMs.
\end{abstract}


\section{Introduction}

Emotion-aware language processing is a cornerstone of human-computer interactions, shaping applications from affective computing and conversational AI to computational psychology and literary analysis \cite{picard2003affective, mohammad2022ethics}. With the rapid advancement of Large Language Models (LLMs), recent research has significantly expanded the capabilities of automated emotional expression generation, providing unprecedented levels of expressive and affective control in generated text \cite{chen2023eccrg, chen2024advancement}. However, effectively leveraging these advances still fundamentally depends on having access to high-quality datasets with sufficiently nuanced emotion annotations and economically diverse language expressions \cite{demszky2020goemotions, saravia2018carer}.

Currently, popular datasets such as the dair-ai/emotion collection cover broadly defined emotional categories including sadness, anger, love, surprise, fear, and joy \cite{saravia2018carer}. While useful and widely embraced by the emotion recognition community, these emotion labels lack granularity, failing to represent wide variations in emotional intensity and complexity frequently encountered in realistic human language scenarios \cite{chen2023eccrg}. Conversely, although fine-grained datasets, such as Google's GoEmotions \cite{demszky2020goemotions}, encompass a richer and more nuanced taxonomy of 27 emotions, they focus primarily on short and informal conversational texts. Consequently, utilizing these datasets to train and evaluate LLM-based controlled emotion generation systems proves challenging, as direct transfer of nuanced conversational emotional annotations to diverse linguistic contexts (e.g., professional writing, poetry, storytelling) often results in unsuitable or unnatural outputs \cite{barnes2023wassa}.

Moreover, emotions are inherently expressed differently across contexts and styles: the same emotional intent may translate with radically different lexical, syntactic, and semantic choices across poetic, conversational, formal, or narrative settings \cite{jurafsky2000speech}. Prior work underscores that ignoring these contextual variations significantly limits the expressiveness and authenticity of computationally generated emotion-laden sentences \cite{john2018disentangled}. Thus, a central research challenge emerges: existing emotion datasets either lack granularity or overlook stylistic contextual diversity, thereby restricting LLMs’ potential to produce genuinely nuanced, contextually accurate emotional content.

Addressing this notable research gap, this paper introduces a systematically constructed ELSA dataset explicitly designed to enrich emotion-controlled text generation frameworks. Our proposed ELSA dataset effectively bridges the granularity of GoEmotion's fine-grained emotional taxonomy with the wide stylistic contextual diversity needed for realistic text generation. By employing mapping strategies linking dair-ai's coarse emotional categories \cite{saravia2018carer} with GoEmotion’s fine-grained categories \cite{demszky2020goemotions}, and through targeted prompt-based augmentation using advanced LLMs, we systematically generate multiple emotionally nuanced rewrites of textual samples across distinct stylistic expressions such as conversational, poetic, formal, and narrative.

Recognizing the importance of rigorous validation, we evaluate our ELSA dataset across established computational metrics, including perplexity for fluency, emotional distinctiveness through embedding variance, lexical diversity (distinct-n and self-BLEU), and style coherence measures \cite{juola2019style}. This careful methodological process guarantees that generated texts exhibit emotional authenticity while preserving stylistic diversity. Thus, our ELSA dataset provides a robust foundation for future NLP research aimed at deeper exploration into emotion-conditioned text generation and style-adaptive affective language modeling.

In summary, the contributions of this paper include: (1) proposing an integrative ELSA dataset bridging existing coarse-grained emotion taxonomies with richer fine-grained emotional labels, (2) introducing nuanced stylistic contextual variation essential for realistic human textual emotion expressions, and (3) providing systematic empirical quality assessments designed to standardize future research benchmarks. Through enhancing emotional and stylistic complexity, the presented dataset promises to significantly advance the capacity of models to accurately replicate human emotional expression across varied communicative situations, thereby setting a valuable precedent for future research in affective computing and computational linguistics.

\section{Related Work}

Emotion recognition and generation in natural language processing (NLP) have undergone significant advancements, primarily driven by the development of annotated emotional datasets and the evolution of generative models capable of producing emotionally nuanced text.

Early approaches to emotion recognition relied heavily on lexicon-based tools such as the Linguistic Inquiry and Word Count (LIWC) and WordNet-Affect, which extended the WordNet database to include affective concepts, facilitating emotion analysis in textual data \cite{strapparava2004wordnet}. While foundational, these lexicon-based methods were limited by manual creation constraints and often struggled to capture the nuanced, context-dependent expressions of emotion.

To address scalability challenges, researchers employed distant supervision techniques, leveraging social media platforms like Twitter to collect large-scale emotion-labeled data. For instance, Wang et al. \cite{wang2012harnessing} automatically created a dataset of approximately 2.5 million tweets by harnessing emotion-related hashtags, enabling broader emotion identification studies. However, these methods introduced substantial labeling noise and often focused on coarse emotional categories, such as those defined by Ekman's six basic emotions such as anger, disgust, fear, happiness, sadness, and surprise \cite{ekman1992basic} or Plutchik's eight primary emotions.

Recent efforts have aimed to enhance the granularity of emotion taxonomies. Notably, Demszky et al. introduced GoEmotions, a dataset comprising 58,000 Reddit comments annotated with 27 distinct emotion categories \cite{demszky2020goemotions}. This dataset provides a more nuanced understanding of emotional expression in text. However, its domain specificity, primarily encompassing informal, conversational-style text from Reddit, may limit its applicability across diverse writing genres.

In parallel, the field of text generation has seen significant progress in controlling stylistic and emotional attributes. Techniques such as controlled text generation have been employed to modulate the emotional tone of generated content. For example, Singh et al. proposed adapting language models to generate affect-driven and topic-focused sentences by incorporating emotion as a prior, allowing control over both the category and intensity of emotion in the generated text \cite{singh2023interpretable}. Similarly, Liu et al. introduced a method for modulating language models with emotions using a technique inspired by computer vision, enabling the generation of context-aware language that embodies diverse emotions \cite{liu2024emollms}. Recent advances in LLM-based augmentation and emotion-controlled generation have opened new avenues for affect-aware text generation \cite{resendiz2023prompt, gandhi2025prompt}. Additionally, models have been developed to condition output on stylistic or emotional goals, bridging the gap between emotion recognition and controlled generation \cite{zheng2022augesc}.

Contemporary models like EmoLLMs and others have begun integrating emotion representations more directly into language modeling \cite{liu2024emollms}. Other studies focus on augmenting low-resource or complex affective phenomena, such as irony, through LLM-powered augmentation \cite{lin2024irony}. More broadly, text augmentation remains a core method to increase emotional data diversity and robustness \cite{singh2023interpretable}. Studies have also attempted to benchmark the emotional expressivity and comprehension of various LLMs \cite{klapach2024emotional}.

Despite these advancements, a notable gap remains in datasets that simultaneously offer fine-grained emotion labels and encompass a broad range of stylistic contexts. Existing resources often lack the stylistic diversity necessary to train models capable of generating emotionally expressive text across various genres, such as formal writing, poetry, storytelling, and conversational language. Addressing this gap is crucial for developing models that can understand and generate text with appropriate emotional and stylistic nuances. Beyond text, emotion conditioning is also extending into other modalities like image generation, enabling broader multimodal affect-aware applications \cite{lin2024diffusion}.

The present study seeks to bridge this gap by introducing ELSA (Emotion and Language Style Alignment) dataset that combines fine-grained emotion annotations with diverse stylistic contexts. By leveraging recent advances in large language model-based text augmentation techniques, we aim to enrich stylized emotional expression for diverse NLP applications, facilitating the development of models capable of generating emotionally and stylistically nuanced text across various domains.

\section{ELSA Dataset Creation Methodology}

To generate a nuanced dataset, an iterative pipeline leveraging advanced LLM-based augmentation was designed:

\subsection{Initial Dataset Preparation}
The initial data is derived from dair-ai/emotion dataset, containing 10,434 annotated text samples across six emotional classes (sadness, anger, love, surprise, fear, joy). First, each dair-ai emotion was mapped systematically to fine-grained GoEmotions subcategories—yielding higher granularity:

\begin{table}[h]
    \centering
    \begin{tabular}{|c|l|}
        \hline
        \textbf{Primary Emotion} & \textbf{Mapped Emotions} \\ \hline
        Sadness & Sadness, Grief, Remorse \\ \hline
        Anger & Anger, Annoyance, Disapproval, Embarrassment \\ \hline
        Love & Love, Admiration, Caring \\ \hline
        Surprise & Surprise, Realization \\ \hline
        Fear & Fear, Nervousness \\ \hline
        Joy & Joy, Excitement, Pride, Gratitude, Amusement \\ \hline
    \end{tabular}
    \caption{Primary emotions and their mapped emotions.}
    \label{tab:emotions}
\end{table}

\subsection{Emotion-conditioned Stylistic Augmentation}
Each original textual instance was expanded through automated LLM-generated augmentation. The original text was fed into an OpenAI GPT o-1 conditioned explicitly with the mapped GoEmotions subcategories to create multiple emotionally varied, yet semantically genuine rewrites. For each emotionally labeled instance, the dataset includes systematically generated stylistic variations: formal, conversational, poetic, and narrative.

\subsection{Quality Control and Validation}

Generated texts were automatically evaluated with off-the-shelf classification models (e.g., GoEmotions classification model) to verify emotional and categorical accuracy. Text embeddings (Sentence-BERT embeddings) were utilized to ensure adequate semantic distance, ensuring novelty and diversity from original sources.

\section{Dataset Metrics and Analysis}

In this section, we present a comprehensive quantitative analysis of our ELSA dataset. To systematically evaluate the quality, emotional distinctiveness, semantic coherence, linguistic diversity, and readability of the generated texts, we employ multiple established metrics. Below, we formally define each metric, summarize their empirical statistics ( Table~\ref{tab:stats_summary}), and discuss our results in detail thereafter.

\subsection{Metric Definitions}

\textbf{Embedding Variance}:  
Given sentence embedding vectors $\mathbf{s}_1, \mathbf{s}_2, \dots, \mathbf{s}_n$ corresponding to $n$ generated samples derived from a common base sentence, the embedding variance is defined as:

\begin{equation}
\sigma^2_{\text{emb}} = \frac{1}{n} \sum_{i=1}^{n} \|\mathbf{s}_i - \bar{\mathbf{s}}\|^2
\end{equation}

where $\bar{\mathbf{s}}$ is the mean embedding vector:

\begin{equation}
\bar{\mathbf{s}} = \frac{1}{n} \sum_{i=1}^{n} \mathbf{s}_i
\end{equation}

\textbf{Average Emotion Distance}:  
For a given emotion embedding representation $\mathbf{e}_{\text{orig}}$ of the original text and generated emotion embeddings $\mathbf{e}_i$, the average emotion distance is computed as the mean cosine distance:

\begin{equation}
D_{\text{emotion}} = \frac{1}{n} \sum_{i=1}^{n} \left(1 - \frac{\mathbf{e}_{\text{orig}} \cdot \mathbf{e}_i}{\|\mathbf{e}_{\text{orig}}\| \|\mathbf{e}_i\|} \right)
\end{equation}

\textbf{Average Readability}:  
The readability score is calculated using the Flesch–Kincaid readability test:

\begin{equation}
R = 206.835 - 1.015 \left(\frac{\text{Total Words}}{\text{Total Sentences}}\right) - 84.6 \left(\frac{\text{Total Syllables}}{\text{Total Words}}\right)
\end{equation}

\textbf{Distinct-$n$}:  
Distinct-$n$ measures lexical diversity as the proportion of unique $n$-grams over all $n$-grams in a corpus. For bigrams ($n=2$), it is computed as:

\begin{equation}
\text{Distinct-2} = \frac{\text{Unique-bigrams-in-corpus}}{\text{Total-bigrams-in-corpus}}
\end{equation}

\textbf{Self-BLEU}:  
Self-BLEU measures textual similarity within a set by calculating the average BLEU score of each sentence against all remaining generated sentences. A lower Self-BLEU implies greater textual diversity:

\begin{equation}
\text{Self-BLEU} = \frac{1}{N} \sum_{i=1}^{N} \text{BLEU}(s_i, \{s_j \mid j \neq i\})
\end{equation}

\textbf{Average Perplexity}:  
Given a probabilistic language model with probabilities $p(w_t \mid w_{<t})$ for tokens $w_t$ in a sentence of length $T$, the perplexity is defined as:

\begin{equation}
PP = \exp\left(-\frac{1}{T} \sum_{t=1}^{T} \log p(w_t \mid w_{<t})\right)
\end{equation}

\textbf{Cosine Similarity}:  
To assess semantic consistency between the original embeddings $\mathbf{s}_{\text{orig}}$ and the generated embeddings $\mathbf{s}_i$, we calculate their cosine similarity, averaged over all generated sentences:

\begin{equation}
\text{Cosine Similarity}_{\text{avg}} = \frac{1}{n} \sum_{i=1}^{n} \frac{\mathbf{s}_{\text{orig}} \cdot \mathbf{s}_i}{\|\mathbf{s}_{\text{orig}}\| \|\mathbf{s}_i\|}
\end{equation}

\subsection{Empirical Metrics Summary}

\begin{table}[h]
    \centering
    \begin{tabular}{|l|c|c|c|c|c|}
        \hline
        \textbf{} & \textbf{Mean} & \textbf{Median} & \textbf{Std} & \textbf{Min} & \textbf{Max} \\ \hline
        \textbf{embedding\_variance} & 0.000867 & 0.000869 & 0.000247 & 0.000179 & 0.001715 \\ \hline
        \textbf{avg\_emotion\_distance} & 0.525050 & 0.527698 & 0.146301 & 0.029889 & 0.890717 \\ \hline
        \textbf{avg\_readability} & 57.671280 & 58.027500 & 10.606247 & 13.865000 & 92.630000 \\ \hline
        \textbf{distinct\_2} & 0.930821 & 0.940594 & 0.052273 & 0.601563 & 1.000000 \\ \hline
        \textbf{self\_bleu} & 0.036381 & 1.369605e-78 & 0.066719 & 8.548274e-232 & 0.525877 \\ \hline
        \textbf{avg\_perplexity} & 67.522142 & 60.261040 & 31.579396 & 17.430520 & 400.349832 \\ \hline
        \textbf{cosine\_sim\_avg} & 0.525050 & 0.527698 & 0.146301 & 0.029889 & 0.890717 \\ \hline
    \end{tabular}
    \caption{Statistical Summary of Various Metrics}
    \label{tab:stats_summary}
\end{table}

\subsection{Discussion and Analysis}

The computed metrics reveal several important insights regarding the ELSA dataset's emotional and stylistic complexity, diversity, readability, and semantic coherence.

\textbf{Embedding Variance (mean=0.000867, std=0.000247)}:
The low variance observed in sentence embeddings indicates highly stable semantic embedding vectors across stylistically diverse text versions derived from the same input base sentences. This stability aligns with expectations, as generated texts differ stylistically and emotionally while maintaining semantic coherence.

\textbf{Average Emotion Distance (mean=0.525, std=0.146)}:
The moderate mean emotional distance suggests that generated variations indeed shift significantly in emotion from the original samples. Thus, our generated ELSA dataset successfully exhibits moderate emotional diversity. However, the presence of high extreme values (max=0.89) implies occasional pronounced emotional shifts, warranting caution when very strict emotional stylistic retention is necessary.

\textbf{Average Readability (mean=57.67, std=10.6)}:
Overall readability scores provide evidence of good linguistic accessibility (Flesch-Kincaid readability scale). Nonetheless, the relatively high standard deviation and wide range of readability scores (minimum 13.86, maximum 92.63) reflect stylistic differences—for instance, markedly complex sentences common in poetic or formal narrative generation may reduce readability, suggesting a potential trade-off between stylistic richness and accessibility.

\textbf{Distinct-2 (mean=0.9308, std=0.052)}:
A mean Distinct-2 score approaching one underscores a high degree of lexical richness across the generated samples. Strong lexical diversity assures minimal repetitiveness, supporting the dataset's effectiveness for diverse generation applications that demand linguistic creativity and richness.

\textbf{Self-BLEU (mean=0.036, std=0.067)}:
The very low Self-BLEU values indicate notable textual novelty and minimal overlap among generated outputs. These low values are particularly advantageous for tasks that value output variety and originality, highlighting dataset quality in ensuring robust diversity.

\textbf{Average Perplexity (mean=67.52, std=31.57)}:
Moderate perplexity levels suggest the generated sentences generally maintain fluency and grammatical structures. However, the significant variations and the presence of high maximum values reflect complexities introduced by stylistic types. Particularly, poetic and highly expressive genres may lead to occasional lower predictability, an issue future research might address to optimize coherence and fluency further.

\textbf{Cosine Similarity (mean=0.525, std=0.146)}:
This value closely mirrors the average emotion distance metric, emphasizing moderate semantic retention relative to original texts while allowing ample room for stylistic and emotional variations.

In summary, these metrics collectively underscore our ELSA dataset's potential, showcasing distinctive emotional and stylistic variations, acceptable readability, high lexical richness and textual diversity, and stable semantic coherence. Such balanced metrics indicate its suitability for various nuanced NLP applications, ranging from emotion modeling and controlled generation to stylistic text generation tasks.

\section{Potential Research Use-cases}

The proposed ELSA dataset opens several valuable avenues for downstream research in affective NLP, LLM fine-tuning, and emotion-conditioned generation. Below, we outline key areas where the ELSA dataset can significantly contribute:

\begin{enumerate}
    \item \textbf{Fine-tuning for Emotionally Nuanced Stylistic Transfer.} Large language models currently excel in generic text generation but struggle with reliably transferring nuanced emotional semantics across diverse stylistic contexts. By fine-tuning these models with the proposed emotionally labeled dataset—containing carefully mapped stylistic variations between conversational, poetic, formal, and narrative texts—researchers can systematically train LLMs to precisely control emotional intensity, subtlety, and semantic appropriateness simultaneously with effective style transfer. This would significantly advance capacities currently limited in state-of-the-art LLM-driven style-adaptive generation systems.

    \item \textbf{Precision-guided Prompt Engineering and Emotional Steering.} Controlling generated emotional outputs of large-scale generative models currently depends primarily on ad-hoc prompt engineering, leading to imprecise emotional or stylistic outcomes. With a comprehensive emotion-stylistic mapping, this dataset facilitates precise fine-tuning that supports systematic prompt-based control of emotional outputs. Consequently, it enables researchers and developers to create structured prompt methodologies—moving beyond empirical trial-and-error prompt crafting—to rigorously guide emotional and style aspects of generated responses, thus substantially advancing controllability and consistency levels currently unachievable through traditional prompting techniques alone \cite{turner2023steering}.

    \item \textbf{Context-aware Emotion Explanation and Interpretability.} Interpretability and explanation of how large language models internally encode subtle emotional distinctions remain open research challenges. The detailed emotional granularity and carefully mapped stylistic contextual variation embedded in the proposed dataset provide uniquely suitable training data to fine-tune LLMs explicitly toward context-aware emotional explainability. Through supervised and semi-supervised fine-tuning, researchers can equip LLMs to generate explicit textual explanations concerning why particular linguistic structures evoke specific emotions within a given context. Such improved interpretability capabilities would not only aid transparency and user-trust but also offer unprecedented insights into linguistic-emotional processing within NLP models, overcoming critical limitations of the current black-box nature of LLMs.

    \item \textbf{Affective Computing and Sentiment Analysis.} Affective computing applications rely fundamentally on high-quality labeled datasets to detect subtle emotion signals within text. Leveraging both fine-grained and coarse-grained emotion labels organized across distinct stylistic contexts, the proposed dataset directly supports improved emotion recognition models. Particularly, its granularity allows for enhanced detection and more precise emotion classification of implicit affective cues within diverse textual settings and contexts, applicable to social media sentiment detection, product review analysis, and textual affective computing in dialogues.

    \item \textbf{Conversational AI and Chatbot Development.} The ability of conversational agents to generate plausible and emotionally expressive responses remains crucial for enhancing user engagement and communicative realism. By offering distinctly annotated emotional expressions across varying conversational styles—formal communication, casual empathetic interactions, and narrative storytelling scenarios—the introduced dataset enables models to maintain stylistically coherent emotional responses. Thereby, it supports the development and improvement of emotionally adaptive dialogue systems capable of responding sensitively across diverse communicative scenarios such as therapeutic conversations, formal negotiations, or emotional social interactions \cite{chen2024cause}.

    \item \textbf{Creative and Literary Narrative Generation.} Emotional intensity and stylistically nuanced language underpin authentic literary and creative storytelling. Prior research stresses that computational narrative systems significantly benefit from data resources explicitly designed to capture emotional and stylistic diversity. This new dataset, featuring systematic variations in emotion and stylistic contexts, provides essential resources enabling narrative generation models to achieve genuine emotional authenticity and literary quality. Positively impacting fields such as automated story writing, digital entertainment, and computational literary analysis, it helps model precise control of emotional affect in generated creative content.

    \item \textbf{Psychological and Social Science Research.} Emotional variations expressed through different linguistic styles provide an important lens for studying human emotions in textual communications and other contextual interactions. Given its granularity in emotional annotations and its stylistically diverse text samples, the proposed dataset affords opportunities for computational psychologists and social scientists aiming to model and predict human emotional reactions across a spectrum of textual stimuli. It can further serve as an empirical basis for increased understanding of the relationships between linguistic style, emotion, and human psychological behaviors and perceptions.

    \item \textbf{Comprehensive Evaluation Benchmarks.} Rigorous model benchmarking remains a fundamental objective in natural language processing. Reliable evaluation frameworks should incorporate metrics for theoretical linguistics assessment, style diversity, and emotional authenticity in generated text. The carefully controlled stylistic and emotional variations offered by this dataset establish ground for robust benchmarks, encompassing measures for fluency (e.g., perplexity), emotional distinctiveness, lexical diversity (e.g., distinct-n, self-BLEU), and contextual sensitivity. Thus, it helps in systematically evaluating model performance, improving transparency, explainability, and practical effectiveness of future emotion-aware NLP systems.
\end{enumerate}

Despite these contributions, careful attention needs to be paid regarding inherent biases within source datasets and possible semantic overlaps or unintended misclassifications during text generation processes. Therefore, we encourage continued research aimed at mitigating such limitations through explicit detection, evaluation, and controlled remediation strategies within downstream applications.

\section{Conclusion}
This study presents a rigorously constructed emotion-conditioned ELSA dataset, explicitly designed to address existing limitations in granular emotional modeling and contextual stylistic variations within NLP. By methodically integrating coarse-grained emotional categories with fine-grained emotional subcategories derived from established datasets (dair-ai and GoEmotions), we generate stylistically diversified emotional expressions via advanced LLM-driven augmentations. Comprehensive computational evaluation demonstrates promising levels of semantic stability, moderate emotional divergence, high linguistic diversity, readability, and fluency, highlighting the ELSA dataset’s practical utility and scientific rigor.

The proposed ELSA dataset opens numerous research pathways conducive to fine-tuning LLMs for emotionally expressive, stylistic text generation tasks previously unattainable, such as nuanced stylistic transfer, controlled emotional prompting, context-driven explainability, and accurate emotional grounding within complex textual contexts. Given these clear benefits, we encourage the NLP community to leverage this resource for advancing emotional control and interpretability with large language models. Future research should focus on addressing inherent biases from source datasets and optimizing generations further for fluency and emotion-stylistic coherence, thereby continuously enhancing the robustness and applicability of emotion-aware NLP solutions.


\begin{thebibliography}{99}

\bibitem{picard2003affective}
Picard, R. W.
Affective Computing: Challenges.
\textit{International Journal of Human-Computer Studies}, 59(1-2), 55--64, 2003.

\bibitem{mohammad2022ethics}
Mohammad, S. M.
Ethics Sheet for Automatic Emotion Recognition and Sentiment Analysis.
\textit{Computational Linguistics}, 48(2), 239--278, 2022.

\bibitem{chen2023eccrg}
Chen, H., Wang, B., Yang, K., and Song, Y.
ECCRG: A Emotion-and Content-Controllable Response Generation Model.
In \textit{Proceedings of the International Conference on Collaborative Computing}, Springer, 2023.

\bibitem{chen2024advancement}
Chen, Y., and Xiao, Y.
Recent Advancement of Emotion Cognition in Large Language Models.
\textit{arXiv preprint arXiv:2409.13354}, 2024.

\bibitem{demszky2020goemotions}
Demszky, D., Movshovitz-Attias, D., Ko, J., Cowen, A., Nemade, G., and Ravi, S.
GoEmotions: A Dataset of Fine-Grained Emotions.
\textit{arXiv preprint arXiv:2005.00547}, 2020.

\bibitem{saravia2018carer}
Saravia, E., Liu, H.-C. T., Huang, Y.-H., Wu, J., and Chen, Y.-S.
CARER: Contextualized Affect Representations for Emotion Recognition.
In \textit{Proceedings of the 2018 Conference on Empirical Methods in Natural Language Processing (EMNLP)}, pp. 3687--3697, 2018.

\bibitem{chen2023vae}
Chen, R., Wang, J., Yu, L.-C., and Zhang, X.
Decoupled Variational Autoencoder with Interactive Attention for Affective Text Generation.
\textit{Engineering Applications of Artificial Intelligence}, 123, 106447, 2023.

\bibitem{barnes2023wassa}
Barnes, J., De Clercq, O., and Klinger, R.
Proceedings of the 13th Workshop on Computational Approaches to Subjectivity, Sentiment, and Social Media Analysis.
\textit{WASSA 2023, Association for Computational Linguistics}, 2023.

\bibitem{jurafsky2000speech}
Jurafsky, D., and Martin, J. H.
Lexicons for Sentiment, Affect, and Connotation.
\textit{Speech and Language Processing: An Introduction to NLP, Computational Linguistics, and Speech Recognition}, 2000.

\bibitem{john2018disentangled}
John, V., Mou, L., Bahuleyan, H., and Vechtomova, O.
Disentangled Representation Learning for Non-Parallel Text Style Transfer.
\textit{arXiv preprint arXiv:1808.04339}, 2018.

\bibitem{juola2019style}
Juola, P., Mikros, G. K., and Vinsick, S.
Correlations and Potential Cross-Linguistic Indicators of Writing Style.
\textit{Journal of Quantitative Linguistics}, 26(2), 146--171, 2019.

\bibitem{strapparava2004wordnet}
Strapparava, C., and Valitutti, A.
WordNet Affect: An Affective Extension of WordNet.
In \textit{Proceedings of LREC}, vol. 4, no. 1083–1086, p. 40, 2004.

\bibitem{wang2012harnessing}
Wang, W., Chen, L., Thirunarayan, K., and Sheth, A. P.
Harnessing Twitter “Big Data” for Automatic Emotion Identification.
In \textit{Proceedings of the 2012 International Conference on Privacy, Security, Risk and Trust, and the 2012 International Conference on Social Computing}, pp. 587--592, IEEE, 2012.

\bibitem{cowen2019mapping}
Cowen, A., Sauter, D., Tracy, J. L., and Keltner, D.
Mapping the Passions: Toward a High-Dimensional Taxonomy of Emotional Experience and Expression.
\textit{Psychological Science in the Public Interest}, 20(1), 69--90, 2019.

\bibitem{cowen2017selfreport}
Cowen, A. S., and Keltner, D.
Self-Report Captures 27 Distinct Categories of Emotion Bridged by Continuous Gradients.
\textit{Proceedings of the National Academy of Sciences}, 114(38), E7900--E7909, 2017.

\bibitem{gandhi2025prompt}
Gandhi, V., and Gandhi, S.
Prompt Sentiment: The Catalyst for LLM Change.
\textit{arXiv preprint arXiv:2503.13510}, 2025.


\bibitem{ekman1992basic}
Ekman, P.
An Argument for Basic Emotions.
\textit{Cognition and Emotion}, 6(3--4), 169--200, 1992.

\bibitem{turner2023steering}
Turner, R., Thorp, J., He, J., and Neubig, G.
Steering Large Language Models via Directional Stimuli.
\textit{arXiv preprint arXiv:2305.14855}, 2023.

\bibitem{chen2024cause}
Chen, C., Wang, J., Feng, Z., Liu, Z., and Li, S.
Cause-and-Effect Prompting for Emotion-Cause Pair Extraction in Conversations.
\textit{arXiv preprint arXiv:2401.10835}, 2024.

\bibitem{liu2024emollms}
Liu, Z., Yang, K., Xie, Q., Zhang, T., and Ananiadou, S.
EMoLLMs: A Series of Emotional Large Language Models and Annotation Tools for Comprehensive Affective Analysis.
\textit{Proceedings of the 30th ACM SIGKDD Conference on Knowledge Discovery and Data Mining}, pp. 5487--5496, 2024.

\bibitem{lin2024irony}
Lin, Y., Xia, Y., and Long, Y.
Augmenting Emotion Features in Irony Detection with Large Language Modeling.
In \textit{Workshop on Chinese Lexical Semantics}, pp. 196--206, Springer Nature Singapore, 2024.

\bibitem{lin2024diffusion}
Lin, Q., Zhang, J., Ong, Y. S., and Zhang, M.
Make Me Happier: Evoking Emotions Through Image Diffusion Models.
\textit{arXiv preprint arXiv:2403.08255}, 2024.

\bibitem{wang2024finegrained}
Wang, K., Jing, Z., Su, Y., and Han, Y.
Large Language Models on Fine-grained Emotion Detection Dataset with Data Augmentation and Transfer Learning.
\textit{arXiv preprint arXiv:2403.06108}, 2024.

\bibitem{klapach2024emotional}
Klapach, N.
The Comparative Emotional Capabilities of Five Popular Large Language Models.
\textit{Critical Debates in Humanities, Science and Global Justice}, 2(1), 2024.

\bibitem{singh2023interpretable}
Singh, C., Askari, A., Caruana, R., and Gao, J.
Augmenting Interpretable Models with Large Language Models During Training.
\textit{Nature Communications}, 14(1), 7913, 2023.

\bibitem{mohammad2023augmentation}
Mohammad, F., Khan, M., Marwat, S. N. K., Jan, N., Gohar, N., Bilal, M., and Al-Rasheed, A.
Text Augmentation-Based Model for Emotion Recognition Using Transformers.
\textit{Computers, Materials \& Continua}, 76(3), 3523--3547, 2023.

\bibitem{resendiz2023prompt}
Resendiz, Y. M., and Klinger, R.
Emotion-Conditioned Text Generation Through Automatic Prompt Optimization.
\textit{arXiv preprint arXiv:2308.04857}, 2023.

\bibitem{li2023stimuli}
Li, C., Wang, J., Zhang, Y., Zhu, K., Hou, W., Lian, J., Luo, F., Yang, Q., and Xie, X.
Large Language Models Understand and Can Be Enhanced by Emotional Stimuli.
\textit{arXiv preprint arXiv:2307.11760}, 2023.

\bibitem{zheng2022augesc}
Zheng, C., Sabour, S., Wen, J., Zhang, Z., and Huang, M.
AugESC: Dialogue Augmentation with Large Language Models for Emotional Support Conversation.
\textit{arXiv preprint arXiv:2202.13047}, 2022.

\bibitem{sun2022personalized}
Sun, T., Wang, C., Song, X., Feng, F., and Nie, L.
Response Generation by Jointly Modeling Personalized Linguistic Styles and Emotions.
\textit{ACM Transactions on Multimedia Computing, Communications, and Applications (TOMM)}, 18(2), 1--20, 2022.


\end{thebibliography}
\end{document}